\documentclass[journal]{IEEEtran} 
\IEEEoverridecommandlockouts
\usepackage{cite}
\usepackage{amsmath,amssymb,amsfonts}
\usepackage{algorithmic}
\usepackage{graphicx}
\usepackage{textcomp}
\usepackage{caption}
\usepackage{subcaption}
\usepackage{xcolor}
\usepackage{subfig}

\newcommand{\gulraiz}[1]{\textcolor{blue}{#1}}
\def\BibTeX{{\rm B\kern-.05em{\sc i\kern-.025em b}\kern-.08em
    T\kern-.1667em\lower.7ex\hbox{E}\kern-.125emX}}

\begin{document}

\title{SARS: A Novel Face and Body Shape and Appearance Aware 3D Reconstruction System extends Morphable Models \\

}

\author{
\IEEEauthorblockN{Gulraiz Khan, Kenneth Y. Wertheim, Kevin Pimbblet, Waqas Ahmed}\\
\IEEEauthorblockA{
\textit{Centre of Excellence for Data Science, Artificial Intelligence and Modelling (DAIM), University of Hull, UK} \\
G.KHAN-2022@hull.ac.uk, K.Y.Wertheim@hull.ac.uk, K.Pimbblet@hull.ac.uk
}
}
\maketitle

\begin{abstract}
Morphable Models (3DMMs) are a type of morphable model that takes 2D images as inputs and recreates the structure and physical appearance of 3D objects, especially human faces and bodies.  3DMM combines identity and expression blendshapes with a basic face mesh to create a detailed 3D model. The variability in the 3D Morphable models can be controlled by tuning diverse parameters. They are high-level image descriptors, such as shape, texture, illumination, and camera parameters. Previous research in 3D human reconstruction concentrated solely on global face structure or geometry, ignoring face semantic features such as age, gender, and facial landmarks characterizing facial boundaries, curves, dips, and wrinkles. In order to accommodate changes in these high-level facial characteristics, this work introduces a shape and appearance-aware 3D reconstruction system (named SARS by us), a c modular pipeline that extracts body and face information from a single image to properly rebuild the 3D model of the human full body. SARS improves accurate semantic surface editing to capture expression-driven facial muscle deformation and appearance changes for different genders, while maintaining identity. SARS has three transformation modules. The first module extracts the aforementioned face semantic features from a 2D image. The second module uses a 3DMM to extract expression blend shape weights from the image and combines them with the face semantic features in a latent space to generate a 3D face model. The third module uses the skinned multi-person linear model (SMPL, not a 3DMM) to extract high-level descriptors of the full body and based on these parameters, generates a 3D body model. Finally, an effective fusion technique is used to replace the 3D body model's face with the 3D face model. In our experiments, the system showed promising results on the 3DHumans dataset and other publicly available datasets.

\end{abstract}


\section{Introduction}
Given the high complexity and variance of human body and faces, image-based 3D human reconstruction is usually considered an open and challenging problem. The variability in the parameters like age, gender, and emotion can lead to inconsistencies in 3D reconstruction results. A few external parameters can also impact the results of 3D modeling, including variable lighting, occlusions and positions. It is imperative to propose a 3D reconstruction engine that can handle all these variations and perform consistently under different environmental circumstances.\\
Despite these challenges, several applications that benefit from realistic 3D modeling have inspired several researchers to pursue research and development in this area. It has a number of applications in the fields of sports, healthcare, augmented and virtual reality, and fashion for virtual try-ons. In these applications, facial details are essential for depicting expressions and age, resulting in enhanced realism of the generated face. For example, animators sometimes wish to add wrinkles or modify lip motions in animated games to improve expressiveness. In multimedia, expression modifications are sometimes needed to satisfy specific requirements. It can also help in crowd behavioral analysis and management for surveillance purposes. The entertainment and gaming industries can use 3D reconstruction to improve 3D object and avatar profiling. \\
Traditionally, researchers employed either a multi-view sequence of frames or a single image with depth information to reconstruct the 3D human \cite{I1, I2, I3, I4}. These systems work in a constrained environment with controlled background and lighting conditions.  Recently, with the advent of deep learning and generative models, considerable studies have been conducted on 3D human reconstruction from a single image \cite{I5, I6, I7}. 3D reconstruction in the wild is still a challenging problem because of multiple factors, including occlusion, face identity and motion, and body-mass ratio. Specific SMPL-based methods handle these complications by deriving insights from image features, such as body parameters (pose, shape, and camera angles), and generating 3D models from these parameters. However, the parametric SMPL has difficulty capturing detailed geometric features because of differences in personal appearance and attire. \\
Traditional methods employ one model to predict the entire body, including the face, which often face challenges due to limited face feature availability and heterogeneous whole-body motion data. \\
In this paper, we propose a modular approach that independently applies 3D modelling algorithms to the face and body separately. This approach captures complex body motions and face variations with more flexibility and precision; it handles occlusion and viewpoint variations for creating 3D models in a more effective way. It comprises a shape and identity-aware 3D morphable model that handles variability in face appearance and a SMPL model for body representation. The face module represents a facial structure effectively by integrating 3DMM blend shapes (shape, texture, and expression blend shapes), a distance map, and a signed distance field with identity features (Age, Gender, and facial keypoints). These features collectively capture small lines, muscle movements, and wrinkles on the underlying 3D facial structure. The SMPL model generates a 3D body model to handle different body poses. A 3D model is a mesh comprising a set of dense vertices and edges, whilst a body pose is a small subset (around 24) of these vertices on the body joint locations.\\
The overall reconstruction system, SARS, uses a fusion module to replace the 3D body model's face with the output of the face module to generate the final whole-body 3D model in a mesh format. Each reconstruction module allows us to fully leverage its state-of-the-art results without compromising accuracy or reliability in practical applications. Our fusion module offers a natural yet practical solution to combine them, seamlessly producing unified whole-body 3d estimation results.\\
In summary, our contributions are manifold, 
\begin{itemize}
    \item The presented approach is an end-to-end solution that simultaneously infers facial identity, geometry, as well as body shape in a single modular framework. Each Module is focused on a specialized task: extracting high-level features, face reconstruction, body reconstruction, and integration, respectively.
    \item Based on our previously proposed MultiHeadCNN, a multi-purpose learning network for predicting age, gender, and facial landmarks. We utilized these high-level features to build a 3D reconstruction model for the face. 
    \item We utilized a fusion module to integrate the 3D face model with a preliminary 3D body model to reconstruct a full-body human avatar from a 2D image.
    \item Overall, the system gains promising accuracy for all targets (Face identity, expression, age and body shape) without reducing efficiency. 
\end{itemize}

This paper's layout is as follows: Section II reviews the current literature on 3D reconstruction of the face and entire body. Section 3 presents a detailed method of the proposed system. Section 4 summarises the experimental setup, describing the training and testing datasets and the results. Finally, the work is concluded, and new research directions are suggested in Section 5.\\

\section{Literature Review}
The last decade has made significant progress in 3D human reconstruction, including the face and body. Different researchers have achieved promising results in finding 3D human poses \cite{L1, L2, L3, L4} that act as baselines for 3D reconstruction. Different methods have been employed for 3D reconstruction of human bodies from two-dimensional images. These include 1) regression-based methods that directly map image features (computer vision features like pixel histograms) to a 3D model (vertices and edges) without deep learning, 2) parametric approaches that augment latent image features with metadata of the object for generating a 3D model, and 3) deep learning-based approaches that do not require metadata, 4) Overlapping with the above three categories, single-image-based approaches focus on addressing the difficulty of inferring 3D structures from limited perspectives, and 5) Overlapping with the above four categories, modular approaches focus on capturing fine-grained details of each body part.
\subsection{Regression-Based Approaches}
A large number of recent methods rely on regression methods for 3D human pose and shape reconstruction.  In order to capture the 3D human structure, these methods \cite{L5,L6, L7, L8, L9} used a range of image features, including 3D body-skeleton positions \cite{L6, L7} and 3D heatmaps \cite{L8, L9}. 3D body-skeleton-based approach turns image features into a skeletal structure, determined by key structural joint positions in 3D space. A joint location is composed of three coordinates x, y and z that represent the shoulder, elbow, knee, fingers and wrist. These intermediate results are then turned into a 3D mesh. Heatmap-based 3D reconstruction produces a probabilistic representation of joint locations in 3D space. A 3D heatmap anticipates a volumetric density map rather than simple individual coordinates, with high-intensity regions denoting the possible locations of individual joints.

\subsection{Parametric Approaches}
Parametric reconstruction models are being used extensively for computing 3D poses and structures of humans. These models take an image and metadata about the imaged human as inputs in order to formulate human body variations with respect to face, body, and hand appearance \cite{L10, L11, L12, L13}. SCAPE \cite{L10} was the first one to capture the human body's shape, structure and pose-dependent variations. \cite{L14} presented SMPL, which comprehends blend shapes (input parameters) for pose in addition to linear blend skinning, enabling mesh deformation and depiction of shape variations.\\
Likewise, specialized models have been designed for individual body parts, such as MANO \cite{L13} for hand variability modeling and a few models for face reconstruction \cite{L14, L15, L16}. Recent achievements have directed the development to full-body parametric models, including Adam \cite{L11}, SMPL-X \cite{L12}, and GHUM/GHUML \cite{L17}, which blend face, hands, and body into a single framework, providing a complete representation of the human body parts.

\subsection{Deep Learning Approaches}
Parametric approaches have been the predominant paradigm for the purpose of 3D human pose and shape reconstruction, but metadata is not always available. Early improvements in this area \cite{L18} focused on learning the parameters required by the SCAPE \cite{L10} model semi-automatically without human intervention. A notable development is the fully automatic method SMPLify proposed by \cite{L19}. SMPLify fits the SMPL model to 2D keypoints using an off-the-shelf keypoint detector \cite{L20} and well-defined prior data (estimates of the required parameters) to direct the optimization process, leading to improved estimates of the parameters and then 3D models. Later advances have extended the SMPLify architecture by adding more cues to the fitting procedure, such as multi-view input \cite{L21}, silhouette information \cite{L22}, and multi-person scenarios \cite{L23}. Recent efforts have demonstrated that models can handle more expressive scenarios in multi-view \cite{L24} and single-view settings \cite{L25,L26}.

\subsection{Single-image-based approaches}
The above-mentioned parametric approaches (subsection B), including SMPL \cite{L14} and Adam \cite{L11}, are now frequently used in recent monocular 3D body reconstruction techniques to depict the full 3D human body. Modern techniques \cite{L27, L28, L29, L30, L31, L32} estimate body parameters from a single RGB image and reconstruct the 3D human body using a deep learning framework (subsection C). Different researchers have proposed non-parametric techniques based on single images; some learn the UV maps \cite{L33} by regression, while others learn the 3D model vertices directly \cite{L34, L35}. A few hybrid methods use deep learning models to produce 2D heatmaps from an image have been proposed. These heatmaps are then used in fitting skeleton models to construct joint angles \cite{L36, L37, L38}. FLAME \cite{L41} and ExpNet \cite{L39} are similar methods that feed a single image to neural networks to compute facial landmarks for 3D face reconstruction \cite{L39,L40,L41}. In \cite{L40}, position maps in UV space are constructed to represent 3D face. 

 \subsection{Modular approaches}
 Some researchers have deployed modular approaches to model different body parts separately \cite{Expose, L12, L26, mrt}. These strategies create 3D models of different body parts with different approaches before integrating them into a whole-body 3D model. The Monocular Total Capture (MTC) method \cite{L26} reconstructs the apparent motion of the body, face, and hand using a 3D deformable mesh model. In this study, the Adam framework \cite{L11} was employed for 3D reconstruction; they developed 3D Part Orientation Fields (POFs), an effective representation that encodes the 3D orientations of body parts from 2D image data. However, the MTC method is computationally costly. Zhou et al. \cite{mrt} employed the 3D Morphable Model \cite{Mofa} for 3D face representation and the SMPL-H model \cite{L12} to reconstruct the whole body, as well as the hands on their own. In this method, the 3DMM parameters including blend shapes, expression, and illumination are computed independently after estimating the 3D postures of the hands and body. Another strategy, ExPose \cite{Expose} eliminated the need for optimisation as they used a deep neural network that jointly predicts the body, hand, and face parameters of the SMPL-X model.\\
To leverage the strengths of these diverse approaches in one single framework, our work adopts a hybrid reconstruction approach that includes elements of parametric, deep learning, single-image-based, and modular approaches.\\

\section{Methodology}
This section elaborates on the overall framework proposed for 3D human body reconstruction. It is called SARS. \\

\begin{figure*}[http]
    \centerline{\includegraphics[width=\textwidth,height=\textheight,keepaspectratio]{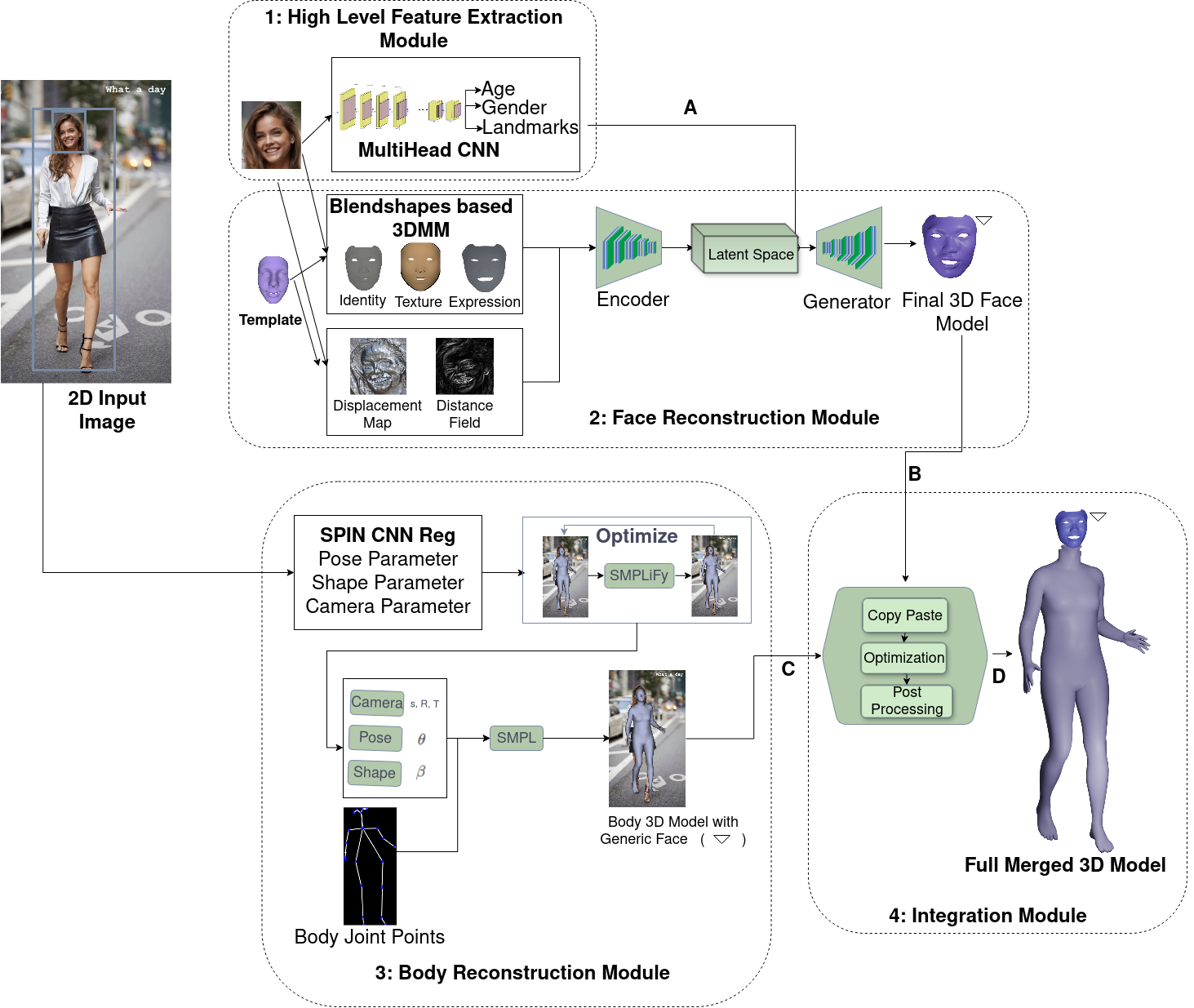}}
    \caption{Full Framework Diagram with three modules highlighted with dotted lines: 1 High Level Feature Extraction, 2 Face Reconstruction Module, 3 Body Reconstruction Module, 4 Integration Module. Each Module's output is represented with capital letters. A shows the values of gender, age, and landmarks; B represents a 3D mesh of the face, having vertices and edges; C shows a 3D mesh of the full body in the representation of dense vertices and edges; D shows the final output of the merged face and body module. $\nabla$ represents the mesh with dense vertices, edges, and triangles}
    \label{fullframework}
\end{figure*}

In a preliminary study \cite{M1}, 1) we developed and validated an extraction module to extract high-level semantic facial features from 2D images, including age, gender, and facial landmarks. In SARS, it works with three other modules. 2) The face module converts a 2D image into 3DMM blend shapes, a displacement map, and a signed distance field, which are encoded in a latent space. This latent space representation is concatenated with the semantic facial features extracted from the same image to generate a refined face model. 3) The body module computes estimates of pose, shape, and camera parameters for the body in the image, improves them with the SMPLiIFY method, and passes the improved parameters and body joints (computed separately) to the SMPL model to generate a 3D body model. Finally, 4) the integration module combines the face and body models to create a seamless and unified whole-body 3D model. The modular structure of SARS allows each component to be optimised independently, potentially resulting in more precise, detailed, controllable and comprehensive outcomes. Figure \ref{fullframework} shows the full connected components of proposed system.\\
\subsection{Face features extraction module}

\begin{figure}[h]
    \centering
  \includegraphics[width=8.7cm, height=8.7cm]{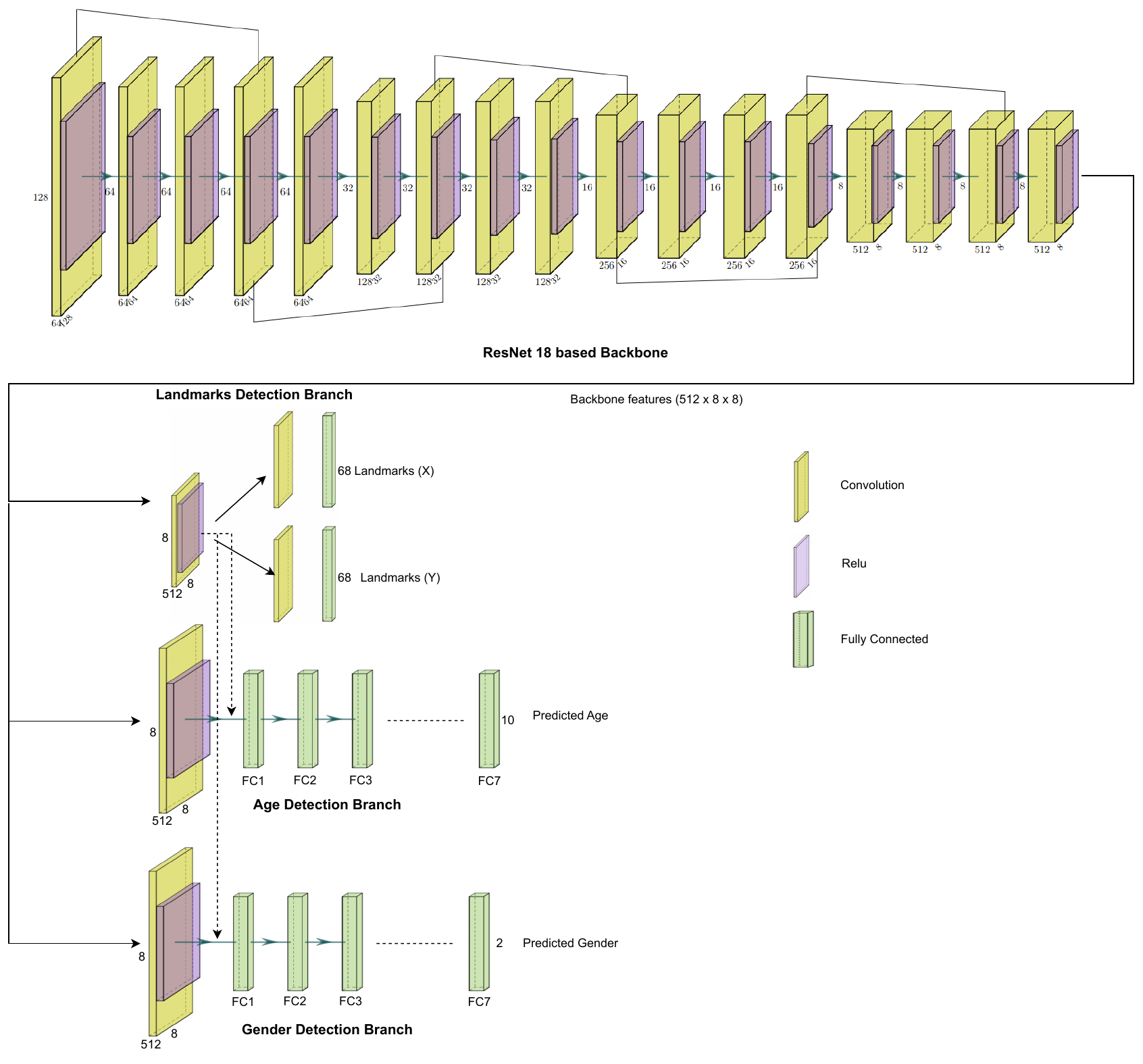}
  \caption{A simplified illustration of the priviously proposed network \cite{M1}. Showcasing the Backbone network generating features with dimensions $518\times8\times8$ and sharing these shared features to three target branches of age, gender, and landmarks.}
  \label{framework}
\end{figure}

In our earlier research \cite{M1}, we proposed a neural network that was primarily designed for multi-task learning (age, gender and facial landmarks), where we designed a single backbone convolutional neural network that computes shared convolution features from the face of an input human image. We exploited the idea that these core facial features can serve as a joint representation for all our target high-level features that eventually eliminates the requirement of dedicated feature extraction networks for each task. The shared features are then passed to three lightweight branches to predict facial landmarks, age, and gender.\\ 
The backbone is built based on residual blocks, which allow handling the vanishing gradient problem by introducing skip connections and enabling gradients to flow more precisely through deep layers. At first, the backbone network takes and processes the input face image, resizing the image to  3 × 256 × 256 and passing it through different convolution and pooling layers and generates a shared deep feature map. The backbone network, comprised of different residual blocks, outputs a feature map of size $512\times8\times8$, which is then shared across all branches. The feature map generated in the backbone network is passed to three dedicated lightweight neural branches: one for computing facial landmarks, the second for age prediction (with 10 age categories), and the third for gender prediction (with binary output). The dedicated branches utilize multiple fully connected layers with only a single convolution layer in each branch to compute the individual target prediction; these fewer layers make them lightweight. This heterogeneous technique enables the network to handle multiple similar tasks effectively while preserving high facial feature extraction and prediction accuracy. The overall architecture of face high-level features is shown in the figure \ref{framework}. \\
\subsection{3D Face Reconstruction Module}
\begin{figure}[http]
    \centering
  \includegraphics[width=10.5cm, height=8.7cm]{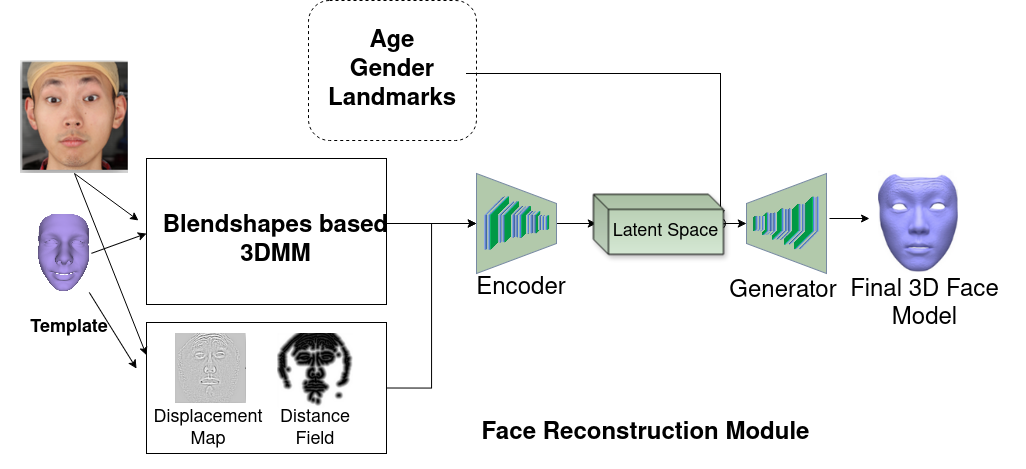}
  \caption{Module 2: Simplified Architecture Diagram of Face Reconstruction based on high-level Semantic Features (Age, Gender and Landmarks)}
  \label{age_exp_edit}
\end{figure}

This module improves the traditional 3DMM using prior parameters (displacement map, signed distance field, and the features extracted by module 1). The process is illustrated in figure \ref{fullframework} and the relevant parts are enlarged in figure \ref{age_exp_edit}.

First, blendshapes are computed from an input image and a template 3D model to represent the shape, identity and expression in the image. In a traditional 3DMM, the blendshapes are used to alter the template 3D model as in equation \ref{eqB}:

\begin{equation}
     F_{3D} = \bar{S} +T+ S_{i}\alpha + S_{e}\beta 
     \label{eqB}
\end{equation}

where $F_{3D}$ represents the target 3D face, $\bar{S}$ is the 3DMM average template shape basis, $S_i$ is the identity blendshape, and $S_e$ is the expression blendshape. Similarly, $\beta$ and $\alpha$ are the vectors of identity and expression coefficients, respectively. The texture $T$ is defined by the linear function in equation \ref{eqT}.

\begin{equation}
    T = \bar{T}+ S_{t}
    \label{eqT}
\end{equation}

where $\bar{T}$ denotes the mean texture and $S_t$ is the blendshape for texture.\\

Improving on this traditional 3DMM, the proposed model accommodates the face's motion and variations in terms of shape and structure representations. This is achieved by incorporating the priors stated above with the blendshapes in a latent space to represent motion and variations, such as expression, wrinkles, and skin smoothness. This approach enhances high-fidelity and identity preservation in 3D face models, capturing both granular and structural facial deformations.\\
In our system, the facial mesh is modeled based on FaceScape 3DMM \cite{Facescape}. This representation is based on the identity parameter ($\alpha$) and the expression ($\beta$). It is important to note that the FaceScape representation concentrates on constructing a precise 3D structure of facial features and does not include attributes of skin texture or lighting effects.\\
To capture small facial muscle movements and structural features, a displacement map is generated from the input image. This map captures subtle surface variations, including wrinkles and small muscle movements. On its own, it enhances the geometric fidelity of a face mesh. A pretrained network that was trained for capturing facial geometry is used to extract a displacement map.
It is computed based on prior work DECA \cite{Feng} that captures scaler offset values for each face pixel along its surface normal. This map enabled adding fine details without modifying the face's original structure.\\
Separately, a signed distance field (SDF) is computed from the input image to model broader facial deformations related to aging or global structural changes. Wrinkle lines are detected from the input image using a pre-trained CNN-based extractor. This makes a dense facial line map depending on texture and shading signals. We extracted signed distance field on the top of detected wrinkle map using Euclidean Distance Transform (EDT) \cite{Linestofacephoto}. On its own, it defines the distance of each vertex in space to the facial contours and guides the global deformation of the mesh, for example, the jaw line. Mathematically, the position of each vertex $p$ is refined according to this equation:

\begin{equation}
    p^{'} = p - \lambda \cdot \text{SDF}(p) \cdot \nabla \text{SDF}(p)
\end{equation}

where $\lambda$ is a step size, $SDF(p)$ indicates the signed distance at point $p$, and $\nabla SDF(p)$ is the gradient of the SDF at point $p$ targeting these components.\\

In SARS, the displacement map and signed distance field are used differently. Together with the blendshapes, they are passed to an encoder. The encoder combines and represents them as a latent vector in a 576-dimensional latent space, which represents granular features, small movements, and global structural variations collectively. This approach allows rich reconstruction and control of dynamic face attributes while maintaining the identity. The encoder models an input image's displacement map $f_d$, the respective distance field $f_f$, and its blendshapes as a latent code $z$ according to the function $z\sim E(input)$, where $input=(f_d,f_f, S_i, S_e, S_t)$. The encoder is based on \cite{swapAE}.\\

In the latent space, the latent vector is concatenated with the other three priors: the high-level extracted features. The purpose is to augment the representation with high-level information about age, gender, and facial landmarks.\\

A generative decoder network takes this concatenated latent vector as its input to reconstruct a detailed 3D face model (facial structure mesh) in an identity-aware and expressive manner. It is based on StyleGAN2 \cite{styleGan2} because of its synthesis quality and adaptability. Again, this mapping can be described in terms of a function, $o_{rec} = G(z)$, where $o_{rec}$ is a set of vertices and edges: the final 3D face model (mesh).\\

A potential application of this module is face editing. By manipulating the high-level priors (age, gender, and facial landmarks), one can control $o_{rec}$ without understanding the structure of the latent space itself.

\subsection{3D Body Reconstruction Module}
This section of the paper describes the 3D human body pose and shape reconstruction module, which optimizes pose and shape parameters with a regression-based technique. A deep learning model called SPIN \cite{SPIN} takes an input image and calculates a set of pose and shape parameters, which are optimized by SMPLify (a regression-based technique) \cite{L19}. The improved parameters are used by a well-established model called SMPL \cite{L14} to generate a 3D mesh.\\

\subsubsection{Initial Parameter Estimation}
In the first step of this module, the pre-trained model SPIN extracts pose $\theta$, shape $\beta$, and camera $\pi$ parameters from an input image. SPIN is a blended framework that combines a convolutional neural network with optimization techniques \cite{SPIN}.\\

\textbf{Shape Parameters ($\beta$)}: They describe how the 6890 vertices in a base mesh are deformed relative to the frame of reference. They indirectly control the height, waist length, volume and proportions of the human body. 
In the SMPL framework, the deviation in body shape is controlled by this low-dimensional parameter vector. We can consider this as a vector of deltas ${\beta} \in \mathbb{R}^{10}$. The overall vertex offset matrix is estimated using a linearer combination of template mesh $\bar{T}$ and  ${\beta} $. Simply consider a template mesh \(\bar{T} \in \mathbb{R}^{6890 \times 3}\) and learned blend shape from shape parameter as \(B_s \in \mathbb{R}^{6890 \times 3 \times 10}\), the shaped mesh ($M_s$)is represented as follow, 

\begin{equation}
    M_s({\beta}) = \bar{T} + B_s ({\beta})
\end{equation}


\textbf{Pose Parameters ($\theta$)}: This is a feature vector with 72 elements. Out of these, 69 elements define the angular displacements relative to the three Cartesian axes (X, Y, and Z) of 23 vertices in the base mesh (representing joints in the human body) that makes overall (23×3) 69 values. The remaining three elements describe how the body as a whole (center of mass) is oriented.\\
Mathmetically, pose-dependent modifications in the mesh ($M_p$) are carried out as follows, 

\begin{equation}
M_p({\theta}) = \sum_{k=1}^{K} \Delta R_k(\boldsymbol{\theta}) \, P_k ,
\end{equation}
Where $\Delta R_k$ shows the rotation of joint $k \in K(23)$ from its original position $R_k$ and $P_K$ is learned blendshape basis.\\
The overall mesh $M_v$ after applying shape and pose parameters is defined nelow,

\begin{equation}
M_v({\theta}, {\beta}) = M_s({\beta}) + M_p({\theta})
\end{equation}

\textbf{Camera Parameters ($\pi$)}: These parameters specify the camera position and angle relative to the human body.\\
\subsubsection{SMPLify}
The three sets of parameters defined above are then optimized by the regression model SMPLify \cite{L19}. It works by iteratively reducing the difference between the 23 3D vertices representing joints and their corresponding 2D keypoints in the input image.\\
\textbf{3D Joint Position Computations}: Using the values of $\beta$ and $\theta$ predicted by SPIN, SMPL computes improved 3D joint positions with a function $J(\beta)$.\\
First, the shape parameters $\beta$ are employed in SMPL to produce a deformed mesh $M_{s}$. This deformed mesh is used to define a joint regressor matrix $\mathcal{J} \in \mathbb{R}^{24 \times 6890}$. The regressor matrix is used to improve the 3D joint positions using a weighted sum of all 6890 vertices. Informally, it means that the vertices close to a joint are given higher weights to compute the joint's position. Then, from the pose parameters $\theta$, a rotation matrix is computed using forward kinematics. The matrix is used to refine the joint positions further by rotating them to form the desired pose. Final vertex positions in deformed mesh ($M$) are represented by the following equation,\\
\begin{equation}
M(\theta, \beta) = \sum_{k=1}^{J} w_{vk} \,
K_k\!\bigl({\theta}, J({\beta})\bigr) M_v({\theta}, {\beta}) 
\end{equation}

In the above equaiton, $w$ is skinning weights and $K$ shows kinematic chain transformations.

\textbf{Projection to 2D}: The 3D joints computed in the previous step are projected onto the 2D image plane using the camera parameter $\pi$, resulting in projected points\\.

The objective of this task is to estimate $L_{joint}$ from the points of the body, with a focus on minimizing the loss of joint detection. The loss for the joint is given by:

\[
L_{\text{joint}} = \sum_i^k \|s_i (j_i - \hat{j}_i)\|,
\]
 
where \(j_i\) denotes the actual 2D joint positions for the \(i\)-th sample, and \(s_i \in \{0, 1\}^K\) represents whether the joint is seeable for kth number, where a value of 1 indicates a visible joint, and 0 indicates hidden, for each of the \(K\) joints.\\

\subsubsection{3D Mesh Generation Using SMPL}
.
The SMPL model is a parametric model that provides 3D depiction of the human body.  This model gets $\theta$ parameters to capture a motion variation in pose and shape parameters $\beta$ to handle variations in apparent shape as input and produces a 3D body mesh. Once we have all three optimized parameters, the SMPL model generates the full 3D mesh of the human body based on the following steps: \\
\paragraph {Shape Blend Shapes}: The shape parameters $\beta$ are applied to a template mesh to reflect individual body shape deviations. This involves finding vertex displacements as a linear combination of blend shapes weighted by $\beta$.\\
\paragraph{Pose Blend Shapes}: The pose parameter $\theta$ is applied to model posture deformations due to joint rotations. \\
\paragraph{Linear Blend Skinning}: Each vertex of the mesh is modified based on the rotations of influencing joints caused by posture change. A weighted summation of transformations from all joints is calculated, where weights are defined for each vertex and represent the effect of each joint on the vertex. It enables smooth, natural movement of a character mesh driven by an underlying skeleton.\\
When the subject changes the rest position, each skeleton joint undergoes a transformation $T_j$, which defines rotation and movement with respect to the rest position. The vertex is refined by incorporating all relevant joint transformations rather than considering only the nearest based on the following,

\begin{equation}
    v_{i}^{'} = \sum_{j=1}^{K} w_{ij} T_j , v_{i}^{0}
\end{equation}

Where $v_{i}^{'}$ is new vertex position, $v_{i}^{0}$ is previous position, $w_{ij}$ represents weight of vertex for each joint and $T_j$ is joint transformation.

We used a dictionary to record the best fit seen for every image in the training set over all epochs in order to improve and speed up training. Every time a new optimized shape is calculated during an iteration, it is assessed against the best output that has been previously recorded. In the event that the new result is better, the record is changed appropriately. The loss on the joints is used to assess the quality of training.
\subsection{Whole-Body Integration Module}

The facial and body reconstruction engine results are combined in our fusion module to create a single representation of the 3D model. Because each module produces a separate reconstruction result, an extra procedure is required to merge the face and body reconstruction results. The most straightforward approach we used was to transfer the relevant mesh from each module separately into the whole-body model.

The facial and body reconstruction modules separately generated high-quality mesh models that should be smoothly combined into a single 3D representation. To achieve this, we first standardized both reconstruction outputs in a shared representation. This method involves normalizing their scales and applying fixed rotations to the vertices. The process also translated the face mesh so that the facial model exactly coincides with the neck region in the body mesh. Our integration system combines three supporting methods: Copy-Paste, Optimization-Based, and Attention-based Processing. The Copy‑Paste method precisely extracts the facial mesh from the face module and overwrites the respective region on the body mesh. While this method is particularly efficient and convenient for real-time systems, it can experience misalignments at the facial boundary, especially near the jawline and neck. 
The reconstructed face mesh ${f_M}$ is combined with ${b_M}$ after removing the face vertices $\mathcal{V}_{face}$ from the body mesh. 
\begin{equation}
    M_c = (M_b \setminus \mathcal{V}_{face}) \cup  M_f
\end{equation}

To remove this irregularity of misalignment, the Optimization‑Based Integration can be employed, as used in PSHuman \cite{M2}. After the copy-paste method, we run optimization that refines the shared parameters (pose, shape, camera) of the combined 3D model. The objective of this function is to reduce seam continuity loss. Seam loss is the vertex-to-vertex difference along the integration boundary and prior terms to maintain reasonable body configurations. To resolve the misalignments, the optimization approach uses a seam loss  ($L_s$) over boundary vertices of the combined mesh $M_c$ and $M_b$. 
\begin{equation}
    L_{\text{s}} = \sum_{i \in \mathcal{S}} \|v_i^{M_c} - v_i^{M_b}\|^2
\end{equation}

Further, joint optimization of the shared parameters $\Theta = \{\beta, \theta, \pi\}$ is achieved by minimizing the combined loss as follows,

\begin{equation}
    L_{\text{c}} =L_{\text{s}} + \lambda_{\text{pose}}\,E_{\text{pose}}(\theta) + \lambda_{\text{shape}}\,E_{\text{shape}}(\beta)
\end{equation}

The gradients from the combined loss update the parameters ($\Theta$) that ultimately create a new refined mesh \cite{M2}. This process notably reduces visual gaps without suffering high computational cost. To further improve visual cohesiveness, especially for the neck, we employ Attentive High-Resolution Processing that refines combined mesh features using body-based attention. This enables the network to learn fine-grained transitions from the face region to the body region. \\
As an essential component of our comprehensive 3D reconstruction pipeline, we incorporate a post-processing stage that smoothly finalizes the merging of facial and body meshes into a single SMPL-X model. After aligning the face and body models using a combination of copy-paste techniques and optimization algorithms, our post-processing technique enhances the 3D model by addressing geometric continuity, filling in holes, and adjusting shading across the interface. This phase employs advanced methods to align surface details from other meshes, refine structural seams, and standardize lighting effects, ensuring a unified and distortion-free reconstruction.

\subsubsection{Alignment Using Geometric Registration}
The alignment of the face and neck subregions is achieved by computing a rigid adaptation that optimally aligns one point set with the other one. This method starts with an initialization based on the principal axes derived from the inertia tensor, which ensures a structurally coherent starting orientation. Following this, the alignment is further improved using an iterative minimization method to reduce point-to-point distances, eventually resulting in a precise overlay at the mesh interface. \\

\subsubsection{Stitching Through Boundary Bridging}
In the processes of stitching, open or uneven edges at the intersection point between the face and body are handled by creating new mesh nodes and edge bridges. These elements are carefully constructed around the border of the gap, often including intermediate vertices to ensure smooth and watertight connectivity.\\

\subsubsection{Vertex Normal Smoothing}
To achieve consistent lighting on the merged mesh, we recomputed vertex normals by taking the mean of neighbouring face normals. This averaging is weighted based on the inner angles of each face, which enriches the continuity of the overall normals and achieves a smoother lighting transition across adjacent areas.\\

\subsubsection{Vertex Normal Smoothing by Nearest Neighbor Mapping}
In order to further minimize visual shading variations, we took a subset of normals from the vertices that represent the neck of the body mesh and transferred these normals to the respective vertices of the face-neck region of the merged mesh. The sampling of body normals is accomplished through a spatial nearest neighbor algorithm, which maintains consistent lighting while assuring that the surface geometry remains unchanged.\\

Figure \ref{fullframework} shows the full framework diagram for the proposed system.\\

\section{Experiments and Evaluation}
Firstly, we present results that our face module outperforms prior state-of-the-art 3D face reconstruction techniques based on MICC Florence dataset using point-to-plane distance, as face priors are used in our system. Parts of the evaluation results were obtained by ablation (explained below), which was performed to examine our design decisions relating to the face module. Secondly, we present evaluation results demonstrating that our whole-body 3D model outperforms other approaches against a commonly used benchmark based on a publicly available dataset.

\subsection{Training Procedures for the Face Module}
In this subsection, we describe the reproducible steps we followed to train the face module with a multi-objective strategy aiming to produce high-quality, semantically equivalent, and structurally precise 3D models of faces.\\

As a reminder of our methodology, this module comprises an encoder (E) and a generator (G). In the forward pass, E encodes the features, which consist of the displacement map ($x_d$), the signed distance field ($x_s$), and three blendshapes, including identity blendshape ($b_i$), expression blendshape ($b_e$), and texture blendshape ($b_t$). These types of features are collectively passed through the encoder to produce a latent space that captures geometric appearance and low-level displacement on the face, including wrinkles, skin deformation, jaw line, and overall face boundary.\\

To make the final 3D face model semantically consistent, the latent vector is then combined with high-level semantic features ($f_{s}$); which contain age, gender, and landmark, and which are computed by the extraction module. The combined feature vector is then passed to G, which produces the final mesh using equation \ref{generatorObj}.

\begin{equation}
    \hat{M}_{g}=G(E(x_d,x_s,b_i,b_e,b_t),f_s)
\label{generatorObj}
\end{equation}

where $\hat{M}_{g}$ shows the final synthesized 3D face mesh, and $f_{s}$ represents the semantic features vector.\\

\subsubsection{Discriminator Objectives}
The training pipeline ends with a discriminator D that follows G. It is trained by backpropagation to produce an equivalent loss function for G, but it is not used for inference. The discriminator D is trained through both adversarial discrimination and semantic categorization tasks. In the adversarial task, it attempts to distinguish between real and generated meshes (adversarial loss). In the categorization task, it attempts to predict the age, gender, and landmarks associated with the 3D facial model (semantic loss).\\

The adversarial loss function is defined as:

\begin{equation}
    L_D^{adv} = - \mathbb{E}_{M}[\log D(M)] - \mathbb{E}_{\hat{M}_{g}}[\log (1 - D(\hat{M}_{g}))] 
\label{advloss}    
\end{equation}

where $L_D^{adv}$ represents adversarial loss and $E$ represents the expectation value of a quantity.\\

The semantic loss, which takes all three categories (age, gender, and facial landmarks) into consideration, is given by the following equation:

\begin{equation}
\begin{split}
L_D^{sem} = & \; \lambda_{age} \, \mathcal{L}_{CE}(D_{age}(M), y_{age}) 
+ \lambda_{gen} \, \mathcal{L}_{CE}(D_{gen}(M), y_{gen}) \\
& + \lambda_{land} \, \mathcal{L}_{CE}(D_{land}(M), y_{land})
\end{split}
\label{semloss}
\end{equation}

where $\lambda$ is a hyperparameter representing the weight of each category, $L_{CE}$ represents cross entropy loss, and $y$ depicts ground truths.

The total loss used to train the discriminator combines equations \ref{advloss} and \ref{semloss}:

\begin{equation}
    L_D = L_D^{adv} + L_D^{sem} 
\label{disloss}    
\end{equation}

\subsubsection{Generator Objectives}
Similar to generative adversarial networks, the negative of $L_D$ in equation \ref{advloss} is used as the first part of the loss function ($L_G^{adv}$) for G in the training pipeline:

\begin{equation}
    L_G^{adv} = - \mathbb{E}_{\hat{M}_{g}}[\log D(\hat{M}_{g})] 
\end{equation}

G has an equivalent but opposite adversarial loss to beat the discriminator in adversarial learning.\\
To ensure that the generated 3D model represents the high-level face features, the following loss function is used:

\begin{equation}
\begin{split}
    L_G^{sem} = \lambda_{age} \, \mathcal{L}_{CE}(D_{age}(\hat{M}_{g}), y_{age}) + \lambda_{gen} \, \mathcal{L}_{CE} \\(D_{gen}(\hat{M}_{g}), y_{gen}) + \lambda_{land} \, \mathcal{L}_{CE}(D_{land}(\hat{M}_{g}), y_{land}) 
\end{split} 
\end{equation}

\subsubsection{Structural Consistency Loss}
For the purpose of maintaining geometric quality and confirming that the constructed 3D mesh is precisely aligned with the distance field of the input, the generator utilized a structural consistency loss. This loss focuses on the significance of maintaining the spatial relationships in the data, enabling a more realistic and cohesive representation of the 3D structure. Structural loss is defined as follows. 

\begin{equation}
L_{\text{struct}} = \mathbb{E} [\lambda_{df} \, \ell_{df}(\hat{M}_{g}, M)]
\end{equation}

In the above equation, $M$ shows the ground-truth mesh,  $\mathbb{E}$ is expectation (mean), and the term $ \ell_{df}$ indicates the amount of the distance-field difference between the reconstructed 3D mesh and the actual 3D mesh.  This technique enables the generator to effectively learn and integrate realistic geometric details that reflect actual 3D facial data.

Finally, the overall objective of the generator can be defined as a combination of semantic, adversarial, and structural loss.

\begin{equation}
  L_G = L_G^{adv} + \lambda_{sem} L_G^{sem} + \lambda_{struct} L_{\text{struct}}   
\end{equation}

we adjust the discriminator to produce expression and age group information of the produced 3D mesh. The discriminator generates a list of vectors $n_{land} + n_{gen} + n_{age}$ based on dataset groups for age, gender and landmarks. Each vector component shows whether a particular mesh belongs to a certain age group, gender, or landmarks orientation.

\subsection{Datasets}
We have performed an evaluation on three recent datasets for evaluating 3D reconstruction. MICC Florence 3D dataset \cite{MiCCAdataset}, 3D Poses in the Wild (3DPW) \cite{E1} and Expressive Hands and Faces (EHF) datasets.

\subsubsection{MICC Florence 3D dataset}

\begin{figure}[http]
    \centerline{\includegraphics[width=9cm,height=\textheight,keepaspectratio]{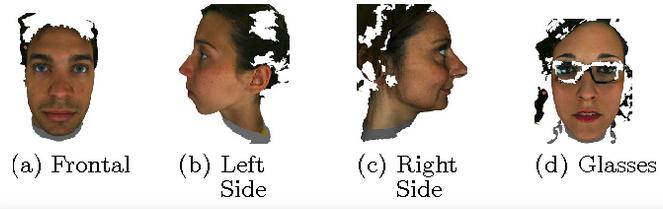}}
    \caption{Sample images from MICC  Florence Benchmark}
    \label{MICCdatasetFig}
\end{figure}

The MICC data \cite{MiCCAdataset} is a significant collection that consists of eight subset datasets, comprising roughly 2,315 individuals under the same FLAME topology.  It concentrates solely on shape geometry and provides registered 3D models with the respective fitted FLAME parameters.  In order to maintain consistency with the original datasets, the data for each subject is arranged into folders with distinct labels.  Although photos are not included in MICA, researchers can find related visual data in primary data.  MICA is a useful tool for 3D face analysis and reconstruction because it is established from representations across various datasets (figure \ref{MICCdatasetFig}). \\

\subsubsection{3D Poses in the Wild (3DPW)}

\begin{figure}[http]
    \centerline{\includegraphics[width=7cm,height=6cm,keepaspectratio]{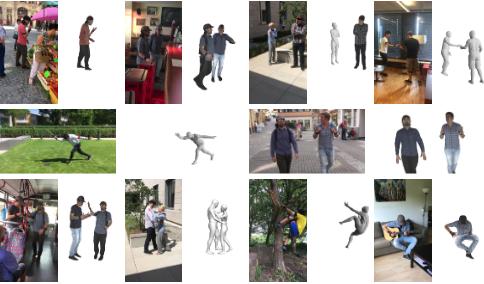}}
    \caption{Sample images from 3D Pose in the Wild Benchmark}
    \label{3DPWFig}
\end{figure}

3D Poses in the Wild (3DPW) \cite{E1} is a demanding dataset for estimating 3D human poses in everyday situations. It comprises 60 video segments, which include over 51,000 frames picturing 7 individuals in 18 various dress designs. This dataset provides a diverse range of  activities such as shopping, sports, hugging, and making selfies. A handheld smartphone camera and 17 IMUs (Inertial Measurement Units) are used for capturing  precise motion in the dataset. Nine to ten IMUs were employed per person to support multi-subject tracking. They used an extra IMU connected to the smartphone for automatic video-IMU synchronization based on clapping motion. Subjects were scanned and annotated with SMPL models, which produced high-quality 3D annotations. Both indoor as well as outdoor recordings are included in the dataset that makes this dataset challenging and comprehensive. Sample images from 3DPW are shown in Figure \ref{3DPWFig}.\\

\subsubsection{Expressive Hands and Faces (EHF) }
\begin{figure}[http]
\centerline{\includegraphics[width=9cm,height=\textheight,keepaspectratio]{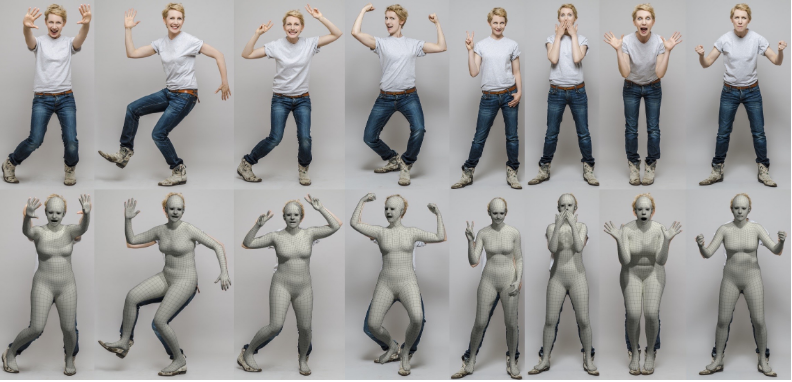}}
    \caption{Sample images from the Expressive Hands and Faces dataset}
    \label{EHF}
\end{figure}

Expressive Hands and Faces (EHF) \cite{E2} is a complex and comprehensive dataset mostly used for evaluating whole-body human pose estimation. This dataset has a particular emphasis on practical hand motions and face reactions. It is made up of 100 RGB photos along with their respective SMPL-X representations.
 The dataset shows each individual participant exhibiting a variety of emotions, complicated hand movements, and dynamic body postures. An expert annotator selected images from the dataset based on the expressiveness of hand and face movements to ensure the quality of the whole dataset. EHF enables researchers to employ a more rigorous vertex-to-vertex (V2V) evaluation metric that captures surface features and fine-grained rotations. Because of this, EHF is a useful benchmark for evaluating whole-body pose estimation models, particularly when it comes to detecting small motion in the hand and face. Samples from EHF dataset are shown in the figure \ref{EHF}.
\subsection{Experimental Results}
The PyTorch open-source software was used to develop the proposed system \cite{E1}. A Nvidia T4 instance with a total of 2560 GPU cores and 16GB of GPU memory was used for training the model. 

\begin{table}[]
\begin{tabular}{lllllll}
\hline
Environment & \multicolumn{2}{c}{Cooperative} & \multicolumn{2}{c}{Indoor} & \multicolumn{2}{c}{Outdoor} \\
 & $\mu$ & $\sigma$ & $\mu$ & $\sigma$ & $\mu$ & $\sigma$ \\ \hline
Anh Tran et al. \cite{L39} & 1.9 & 0.3 & 2 & 0.3 & 1.9 & 0.2 \\
J. Booth et al. \cite{tab1R2} & 1.8 & 0.3 & 1.9 & 0.2 & 1.6 & 0.2 \\
K. Genova et al. \cite{tab1R3} & 1.5 & 0.1 & 1.5 & 0.1 & 1.5 & 0.11 \\
Ours & 0.9 & 0.1 & 0.9 & 0.1 & 0.9 & 0.1 \\ \hline
\end{tabular}
\caption{Comparison of face reconstruction methods on the MICC Florence 3D dataset, evaluated using the mean ($\mu$) and standard deviation ($\sigma$) of the average symmetric point-to-plane distance (lower is better). The evaluation is based on single-frame reconstructions, with meshes captured around the nose area. ICP dense allighnment is used to allign predicted mesh with ground truth, in line with prior work.}
\label{Table1}
\end{table}
\subsubsection{Face Reconstruction Evaluation}
Table \ref{Table1} outlines the performance of our proposed 3D face reconstruction module using the MICC Florence 3D dataset (MICC) \cite{MiCCAdataset}. This dataset contains 3D scans of 53 people in three different, varying complex environments: cooperative, indoor, and outdoor. Instead of processing several frames as done in previous works like \cite{tab1R3,Tuan}, we streamline the pipeline by concentrating on single-frame processing, guided by high-level facial feature priors. By doing so, multi-image support is no longer necessary, and our method can provide accurate reconstructions with just one video frame.\\
We adopted FaceScape topology because of its exceptional resolution in facial structure and detailed expression abilities. It is not feasible to compare FLAME and FaceScape directly at parameter level as they differ in mesh structure, number of vertices, and parametric representation. We employed rigid Iterative Closest Point (ICP) \cite{ICP} for allignment purpose and eliminate translation, rotation, and scale disparities.\\
We restricted each reconstructed mesh to a small area near the center point of the nose to allow us to concentrate on the facial internal structure. In accordance with the suggestions made in \cite{Tuan}, this cropping provides a benchmark evaluation area. We employed the Iterative Closest Point (ICP) algorithm \cite{ICP}, a proven technique for achieving dense alignment by eliminating geometric gaps between corresponding points on the two surfaces, to compare the predicted meshes with the ground truth. As utilized in previous studies \cite{tab1R3}, the average symmetric point-to-plane distance, a metric that measures the deviation between the reconstructed and true mesh surfaces, is used to quantify the accuracy of the shape reconstruction. This approach enables us to evaluate the model's capacity to reconstruct the fine facial details with accuracy and consistency.

As depicted in Table \ref{Table1}, our model significantly surpasses existing approaches, lowering the normalized point-to-plane error by 40\%. Furthermore, our system exhibits robust performance across all testing environments with minimal deviation in the reconstruction output in all environments (indoor, outdoor and Cooperative), indicating its reliability in various scenarios (different poses, expressions, lightening conditions and light angles).\\

\begin{figure}[htbp]
\centering
\setlength{\tabcolsep}{1pt} 
\renewcommand{\arraystretch}{0.1} 

\begin{tabular}{c@{\hspace{4pt}}ccc}
\raisebox{2\height}{\rotatebox{90}{\textbf{Input}}} &
\includegraphics[width=0.15\textwidth,height=0.15\textwidth]{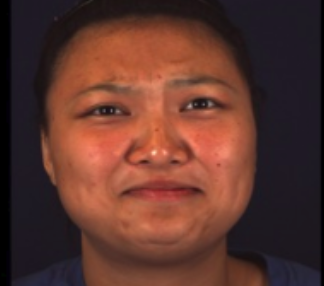} &
\includegraphics[width=0.15\textwidth,height=0.15\textwidth]{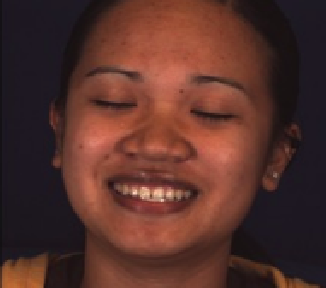} &
\includegraphics[width=0.15\textwidth,height=0.15\textwidth]{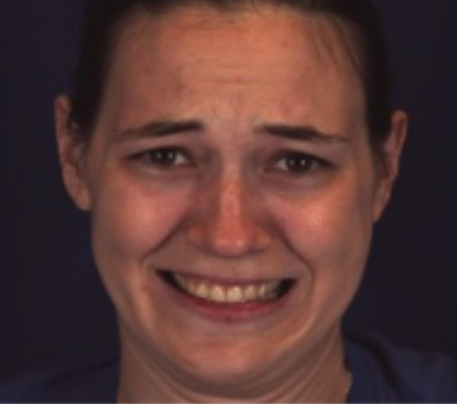} \\[3pt]

\raisebox{0.5\height}{\rotatebox{90}{\textbf{3DDFA \cite{3DDFA}}}} &
\includegraphics[width=0.15\textwidth,height=0.15\textwidth]{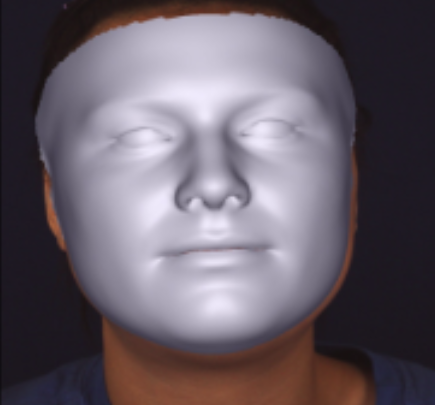} &
\includegraphics[width=0.15\textwidth,height=0.15\textwidth]{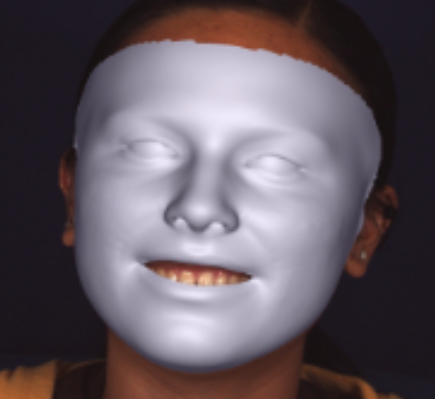} &
\includegraphics[width=0.15\textwidth,height=0.15\textwidth]{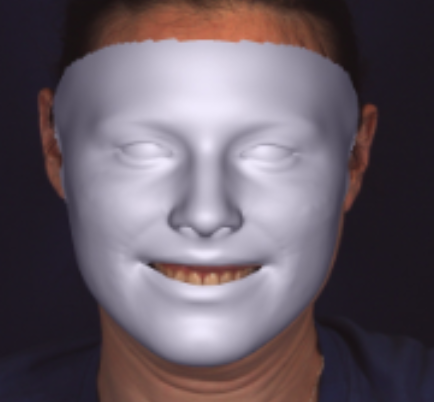} \\[3pt]

\raisebox{0.15\height}{\rotatebox{90}{\textbf{Bai et. al \cite{Bai}}}} &
\includegraphics[width=0.15\textwidth,height=0.15\textwidth]{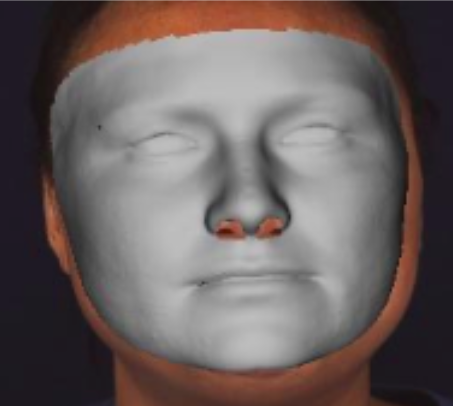} &
\includegraphics[width=0.15\textwidth,height=0.15\textwidth]{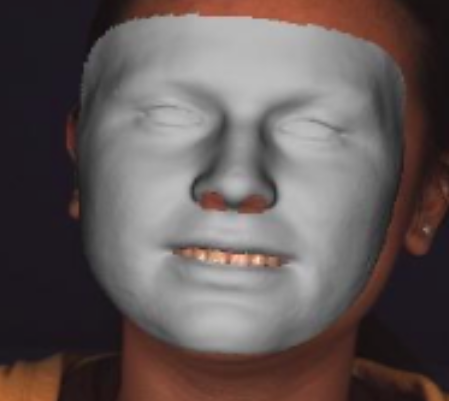} &
\includegraphics[width=0.15\textwidth,height=0.15\textwidth]{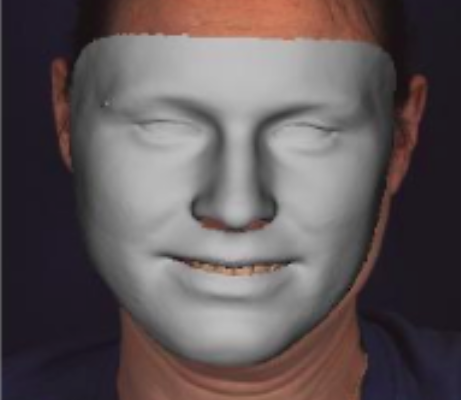} \\[3pt]

\raisebox{0.15\height}{\rotatebox{90}{\textbf{Feng et. al \cite{Feng}}}} &
\includegraphics[width=0.15\textwidth,height=0.15\textwidth]{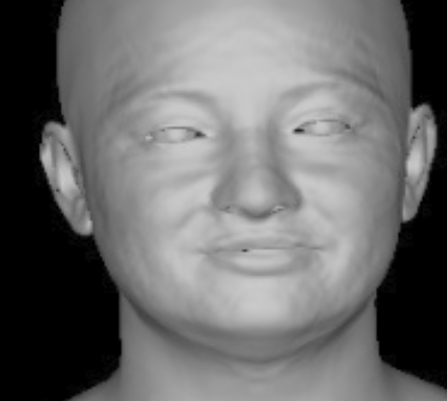} &
\includegraphics[width=0.15\textwidth,height=0.15\textwidth]{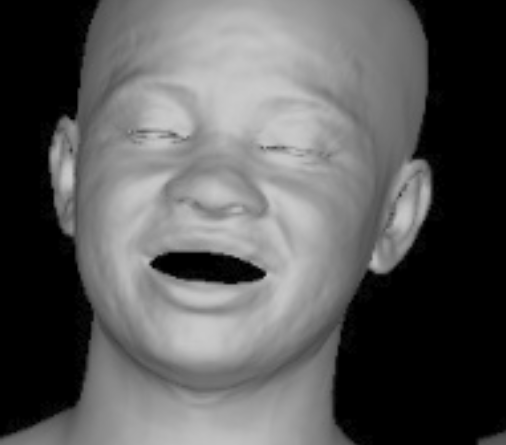} &
\includegraphics[width=0.15\textwidth,height=0.15\textwidth]{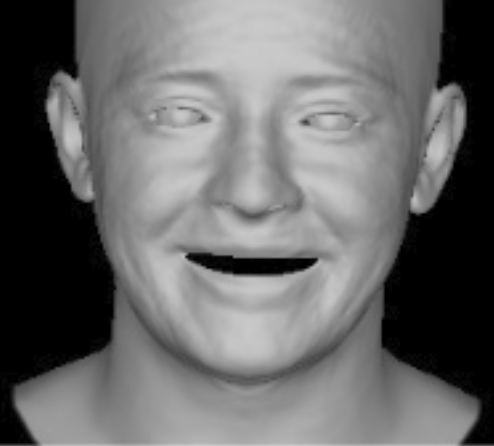} \\[3pt]

\raisebox{0.15\height}{\rotatebox{90}{\textbf{AU-Aware \cite{AU-Aware}}}} &
\includegraphics[width=0.15\textwidth,height=0.15\textwidth]{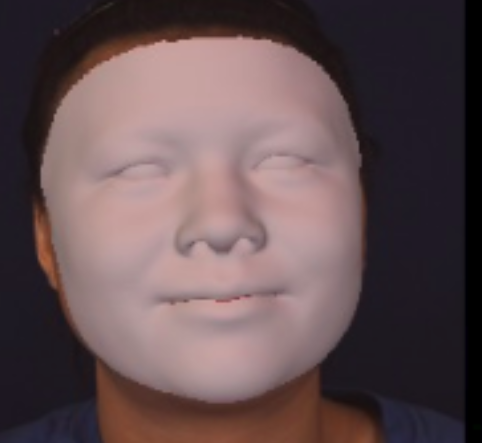} &
\includegraphics[width=0.15\textwidth,height=0.15\textwidth]{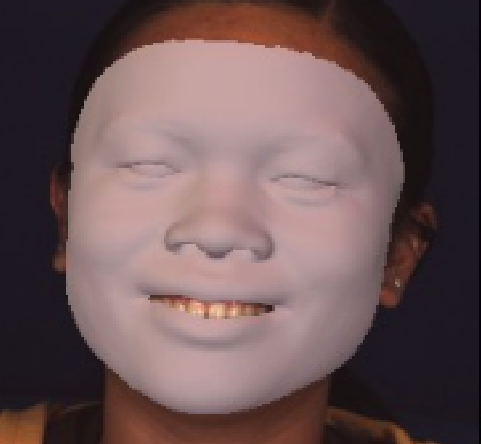} &
\includegraphics[width=0.15\textwidth,height=0.15\textwidth]{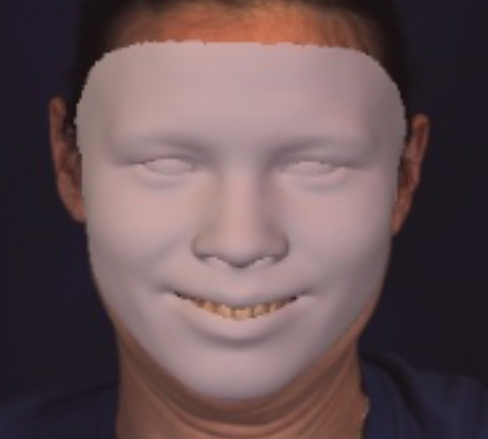} \\[3pt]

\raisebox{2\height}{\rotatebox{90}{\textbf{Our}}} &
\includegraphics[width=0.15\textwidth,height=0.15\textwidth]{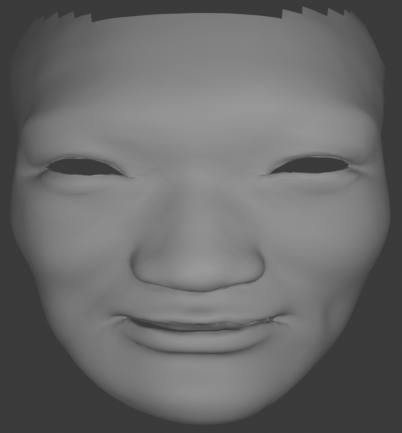} &
\includegraphics[width=0.15\textwidth,height=0.15\textwidth]{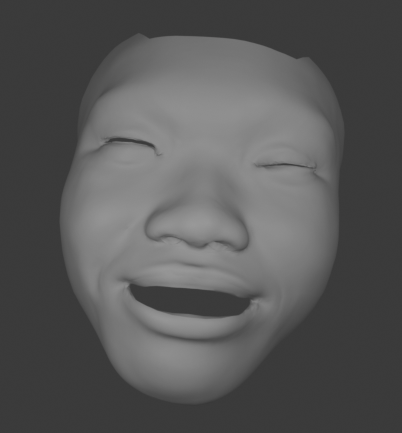} &
\includegraphics[width=0.15\textwidth,height=0.15\textwidth]{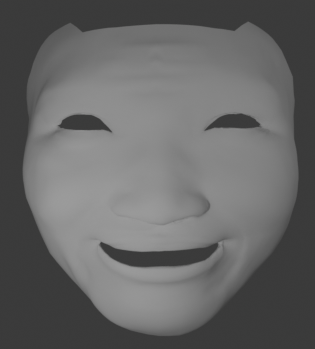} 
\end{tabular}
\caption{Visual comparison of faces reconstructed by our model and those generated by existing state-of-the-art methods}
\label{ResultsComparison}
\end{figure}
A comparison of our engine results with those of previous face reconstruction research \cite{3DDFA, Bai, Feng, AU-Aware}, is illustrated in Figure \ref{ResultsComparison}. Following the input photos in the first row, the next four rows show a comparison of our reconstruction results with the techniques suggested in \cite{3DDFA, Bai, Feng, AU-Aware}.  Our system generates identity and pose consistent 3D face that capture detailed wrinkle lines and expression-driven texture distortions. While some methods, such as \cite{Feng}, achieve adequate surface definition and recover sharper details (e.g., eyebrow contours), our method focus more on identity awareness and geometric deformations across different expressions and age. Our reconstructions effectively capture different facial features, including eyebrow contours, wrinkles, dimples, and the angles of the mouth and eyes.\\
We Performed ablation study of our proposed face reconstruction engine to understand how the full concatenated system performs better than previous variations. As in our approach, we introduced different types of losses, including identity loss $L_{id}$ based on a previously developed face recognition network \cite{Ganfit}, distance field loss $\ell_{df}$, structure loss $L_{struct}$, and semantic loss $L_{sem}$. We tested the system with different variations by removing each type of loss one by one. Table \ref{Albation_Study} shows the Ablation Study with different variations of loss functions.

\begin{table}[]
\centering
\begin{tabular}{|c|c|c|}
\hline

Method & Reconstruction  \\ \hline
Removed $L_{id}$ & 0.1514  \\ \hline
Removed $\ell_{df}$ & 0.1323  \\ \hline
Removed $L_{sem}$ (age,exp) & 0.1303 \\ \hline
Removed  $L_{struct}$& 0.1434  \\ \hline
Combined & 0.1022  \\ \hline
\end{tabular}
\caption{Ablation Study: Learned perceptual image patch similarity error (LPIPS) on MICC data. The system is evaluated with different loss combinations}
\label{Albation_Study}
\end{table}

\begin{table}[http]
\centering
\begin{tabular}{|l|l|l|}
\hline
 & MPJPE & PA-MPJPE \\ \hline
Human Mesh Recover{[}33{]} & 130 & 81.3 \\ \hline
SMPL oPtimization IN the loop {[}37{]} & 96.9 & 59.2 \\ \hline
Exemplar Fine-Tuning {[}30{]} & 92.3 & 54.2 \\ \hline
ExPose {[}18{]} & 93.4 & 60.7 \\ \hline
Our & 93.3 & 53 \\ \hline
\end{tabular}
\caption{Comparison of Reconstruction Results on 3DPW dataset using Mean Per-Joint Position Error (MPJPE) and Procrustes-Aligned MPJPE (PA-MPJPE) distance measures.}
\label{bodyComparison}
\end{table}

\subsubsection{Body Reconstruction Evaluation}
This subsection primarily demonstrates the evaluation results for body reconstruction. We have employed two recent datasets for evaluation 3DPW \cite{E1} body pose dataset and EHF \cite{E2} for testing and training. 
\\
Our system is primarily based on SMPL supervision or representation. We employed  3DPW training set for training for body module as this dataset provides detailed mesh representation for body pose and shape acquired in real-world outdoor conditions. During our training, we use the direct SMPL representation to model the body. This method enabled us to have direct and consistent learning and eliminated the need for complicated model conversions.\\
We also employed the Vertex-to-Vertex (V2V) difference, which is the average divergence between the vertices of the predicted and ground-truth meshes. This method is primarily used when the predicted mesh and ground-truth mesh have the same structure. \\
Using the 3DPW test dataset, which includes SMPL ground-truth meshes, we first evaluated our methodology. Although it lacks details of face structure, this dataset is ideal for comparing full-body reconstruction approaches with SOTA literature including HMR \cite{I5} and SPIN \cite{SPIN}. The proposed system performs better than previous techniques: ExPose, HMR, and SPIN model, comparison results are displayed in \gulraiz{Table \ref{bodyComparison}}. This illustrates how our body reconstruction module provides a realistic and reliable 3D model, allowing for more intricate facial, body and combined reconstructions. The proposed body module can be improved using a modular approach and can still be used in full human reconstruction. 

\begin{table}[http]
\centering
\captionsetup{width=0.5\textwidth}
\begin{tabular}{|l|c|c|c|c|}
\hline
\textbf{\begin{tabular}[c]{@{}l@{}}Part \&\\ Metrics\end{tabular}} & \textbf{\begin{tabular}[c]{@{}c@{}}SMPLify-X \\ \cite{L12}\end{tabular}} & \textbf{\begin{tabular}[c]{@{}c@{}}ExPose \\ \cite{Expose}\end{tabular}} & \textbf{\begin{tabular}[c]{@{}c@{}}FrankMocap \\ \cite{Frankmocap}\end{tabular}} & \textbf{Att-Opt} \\ \hline
\textbf{W-Body V2V} & 146.2 & 81.9 & \textbf{63.5} & \textbf{62.1} \\ \hline
\textbf{W-Body PA-V2V} & 68.0 & 57.4 & \textbf{54.7} & \textbf{53.8} \\ \hline
\textbf{Body V2V} & 231.3 & 116.4 & \textbf{83.7} & 78.3 \\ \hline
\textbf{Body PA-V2V} & 75.4 & 55.1 & \textbf{52.7} & 53.5 \\ \hline
\textbf{Face V2V} & 23.6 & \textbf{19.5} & 25.4 & 18.8 \\ \hline
\textbf{Face PA-V2V} & 8.4 & 5.6 & \textbf{5.4} & \textbf{5.5} \\ \hline
\end{tabular}
\caption{Comparison of Reconstruction Results on EHF dataset using V2V and PA-V2V distance measures with our attention-optimization (Att-Opt) approach in last column. The Comparison is based on SMPL-X representation fitted using SMPLify-X \cite{L19} to handle original SMPL representation.}
\label{fullbodyComparison}
\end{table}

\subsubsection{Whole-Body Integration Module Evaluation}

Using the EHF dataset \cite{E2}, we statistically evaluated our methodology against other whole body approaches from the literature based on SMPL-X, including SMPLify-X \cite{L19} and ExPose \cite{Expose} for the whole body module. We have used the vertex-to-vertex difference and the Procrustes estimation of V2V to evaluate the variation in the shapes of the results. \\
Vertex-to-vertex distance (V2V) and Procrustes estimation vertex-to-vertex distance (PA-V2V) are used as metrics for the evaluation.  The comparison results in table \ref{fullbodyComparison} clearly show that our system outperforms previous systems in terms of the realism of body parts. \\
We evaluated three different integration methods for whole-body reconstruction: Copy-Paste, Optimization-Based Integration, and Attentive High-Resolution Processing. The Copy directly sub-modules offers a light and speedy alternative, making the system better for real-time applications, but it sometimes missalligh hand part. \\
 The optimization-based approach optimizes all parameters for each part to be included in the whole body, thereby reducing the objective function cost. This improves the overall misalignment with little extra computation. Other than that, the full-body attention mechanism utilizes full-resolution photos and focuses on the body parts only. Our results show that while each method has benefits, the best results are gained by combining the optimization approach with the attention mechanism. This fusion offers realism and robustness by remarkably improving the overall quality of the reconstruction. Table \ref{Albationtable} presents an ablation study based on the above integration approaches.

\begin{table}[htt]
\begin{tabular}{|l|cc|ll|ll|}
\hline
\textbf{Methods →} & \multicolumn{2}{c|}{Copy} & \multicolumn{2}{c|}{Simple-OPT} & \multicolumn{2}{c|}{Att-OPT} \\ \hline
\multicolumn{1}{|l|}{\textbf{}} & \multicolumn{1}{c|}{V2V} & PA-V2V & \multicolumn{1}{c|}{V2V} & \multicolumn{1}{c|}{PA-V2V} & \multicolumn{1}{c|}{V2V} & \multicolumn{1}{c|}{PA-V2V} \\ \hline
\multicolumn{1}{|l|}{\textbf{Whole-Body}} & \multicolumn{1}{c|}{73.2} & 58.8 & \multicolumn{1}{l|}{66.4} & 56.1 & \multicolumn{1}{l|}{\textbf{62.1}} & \textbf{53.8} \\ \hline
\multicolumn{1}{|l|}{\textbf{Body}} & \multicolumn{1}{c|}{\textbf{78.6}} & \textbf{54.7} & \multicolumn{1}{l|}{77.9} & 53.4 & \multicolumn{1}{l|}{78.3} & 53.5 \\ \hline
\multicolumn{1}{|l|}{\textbf{Face}} & \multicolumn{1}{c|}{31.2} & 6.3 & \multicolumn{1}{l|}{20.8} & 5.3 & \multicolumn{1}{l|}{18.8} & \textbf{5.5} \\ \hline
\end{tabular}
\caption{Albation Study of proposed system with respect to different integration approaches}
\label{Albationtable}
\end{table}

\section{Conclusion and Future Work}
In this paper, we have proposed a comprehensive framework for the reconstruction of a high-fidelity 3D whole body mesh and enabling identity awareness based on multiple semantic and geometric face priors. Different geometric face priors include displacement map, signed distance field, and blendshapes, and semantic face priors include high-level face semantics features such as age, gender, and face landmarks. The above attributes are combined to achieve higher structural accuracy while maintaining expression and identity. We have also introduced a multi-task discriminator to create a realistic face mesh by validating the semantic categories of the generated mesh. This improvement allows us to reflect semantic features in a more optimised way. \\

Future extension of the proposed system can focus on enhancing the control over modifying the semantic expressions of the reconstructed mesh by changing the high-level features manually. Another promising direction to extend this work is to encompass body volume and weight in body reconstruction. These integrations would not only improve the facial details but would also enhance the morphological attributes of the human body. Moreover, another future research route could be integrating the Implicit Neural Radiance Field.  Finally, examining multidisciplinary capabilities and optimizing the time of the system for real-time applicability would be beneficial for augmented and virtual reality.

\end{document}